%% file: main.tex
\documentclass[10pt, a4paper]{article}
\usepackage[final]{lrec2026}

\usepackage{forest}
\usepackage{booktabs}
\usepackage{CJKutf8}
\newcommand{\ja}[1]{\begin{CJK*}{UTF8}{goth}#1\end{CJK*}}
\usepackage{inconsolata}
\usepackage{amsmath}
\usepackage{multirow}
\usepackage{expex}
\usepackage{tipa}

\newcommand{\totaldatasethours}[0]{6.33}
\newcommand{\bestromajicer}[0]{14.8}

\title{Automatic Speech Recognition for Documenting Endangered Languages: Case Study of Ikema Miyakoan}

\name{Chihiro Taguchi$^1$, Yukinori Takubo$^2$, David Chiang$^1$} 

\address{$^1$University of Notre Dame, $^2$National Institute for Japanese Language and Linguistics \\
         $^1$Notre Dame, IN, USA, $^2$Tokyo, Japan \\
         ctaguchi@nd.edu, ytakubo@ninjal.ac.jp, dchiang@nd.edu}

\input{src/0_abstract}

\begin{document}

\maketitleabstract

\input{src/1_introduction}
\input{src/2_related}
\input{src/3_dataset}
\input{src/4_experiments}
\input{src/5_analysis}
\input{src/6_conclusion}

\section{Ethics statements}
This research is grounded in close collaboration with the Ikema-speaking community and adheres to ethical standards for language documentation and computational research on endangered languages.
All speakers participated voluntarily with informed consent, and their privacy and data rights were respected throughout data collection and processing.
The speech recordings are used strictly for research and documentation purposes, and any future public release of the dataset will follow community consultation and appropriate licensing to ensure responsible use.
Potentially personally-identifiable names that appear in our datasets are annotated with the tags \texttt{<name>...</name>} so that such samples can be removed automatically.

We recognize that the deployment of language technologies in small and vulnerable communities entails complex ethical and social implications.
While ASR systems can effectively accelerate transcription and support revitalization efforts, they may also raise concerns regarding data ownership, representation, and the potential misuse of linguistic data.
To mitigate these risks, we emphasize transparent documentation of data ownership and licensing, in close collaboration with the community and through shared decision-making.

Another relevant important consideration is that developing an ASR system for a traditionally unwritten language may privilege a particular writing system that is not necessarily widely supported within the community.
While the constructed dataset uses a \textit{kana} (\textit{hiragana}) orthography, some speakers in fact find this orthography difficult to read.
This study, as well as the dataset and model developed in this work, does not intend to impose the use of the \textit{kana} orthography on the community.

More broadly, this work contributes to the equitable development of language technologies by extending computational research beyond high-resource languages. By developing resources and methodologies for Ikema, we aim to promote linguistic diversity and inclusivity within NLP, demonstrating that advances in speech technology can serve both scientific and community-centered goals when guided by ethical collaboration and respect for speaker agency.

\section{Acknowledgments}
The material was based on work supported in part by the US National Science Foundation under Grant Number BCS-2109709 and IIS-2137396.
We also thank the Ikema Miyakoan native speakers who helped the authors collect the data and review the model output.

\section{Appendix}
Table~\ref{tab:dataset-full} lists the field recordings used in the Field dataset, as well as their basic statistics.

\input{table/dataset}

\section{Bibliographical References}\label{sec:reference}

\bibliographystyle{lrec2026-natbib}
\bibliography{references}

\end{document}

%% file: src/0_abstract.tex
\abstract{
Language endangerment poses a major challenge to linguistic diversity worldwide, and technological advances have opened new avenues for documentation and revitalization.
Among these, automatic speech recognition (ASR) has shown increasing potential to assist in the transcription of endangered language data.
This study focuses on Ikema, a severely endangered Ryukyuan language spoken in Okinawa, Japan, with approximately 1,300 remaining speakers, most of whom are over 60 years old.
We present an ongoing effort to develop an ASR system for Ikema based on field recordings. 
Specifically, we
(1) construct a {\totaldatasethours}-hour speech corpus from field recordings,
(2) train an ASR model that achieves a character error rate as low as 15\%, and
(3) evaluate the impact of ASR assistance on the efficiency of speech transcription.
Our results demonstrate that ASR integration can substantially reduce transcription time and cognitive load, offering a practical pathway toward scalable, technology-supported documentation of endangered languages.
 \\ \newline \Keywords{language documentation, automatic speech recognition, language resources} }

%% file: src/1_introduction.tex
\section{Introduction}
Language endangerment is a pressing global issue, with thousands of languages at risk of disappearing within the coming decades.
Recent advancements in language technologies have opened up new opportunities for language documentation, offering computational tools that can assist researchers in preserving linguistic data more efficiently.
In particular, automatic speech recognition (ASR) has emerged as a promising technique to support the transcription of spoken data from endangered languages.

Ikema Miyakoan (Ikema henceforth), a highly endangered Ryukyuan language spoken on the Miyako Islands in Japan, represents one such case.
With an estimated speaker population of approximately 1,300 individuals, most of whom are over 60 years old, the language faces severe risk of extinction \citep{nakama-2025-ikema}.
In this context, leveraging speech technologies for Ikema is both urgent and valuable for documentation and revitalization efforts.

This paper reports on our ongoing work to develop a speech dataset and an ASR system for Ikema as part of a broader language documentation project.
We describe the construction of a {\totaldatasethours}-hour speech dataset collected from field recordings, pronounced dictionary entries, and audio books.
Then, we report the training of an ASR model that achieves a character error rate (CER) as low as 14.8\%.
The trained model is then integrated into the linguistic annotation software, and we evaluate how much ASR can facilitate the transcription process.
Our findings indicate that ASR-assisted transcription not only speeds up annotation but also reduces the cognitive burden on annotators, highlighting the practical benefits of integrating ASR into endangered language documentation workflows.

The key contributions of this study are:
\begin{itemize}
    \item The construction of the first speech dataset for Ikema;
    \item The development of the first ASR model for Ikema with a CER as low as {\bestromajicer}\%;
    \item The integration of the ASR model into the linguistic annotation tool (ELAN);
    \item Presenting the positive evidence for the potential of ASR-assisted transcription in language documentation.
\end{itemize}
The dataset, the model, and the experimental code developed in this study are publicly released.\footnote{\url{https://github.com/ctaguchi/ikema_asr}}

%% file: src/2_related.tex
\section{Related Work}
\subsection{Language}
\begin{figure}
    \centering
    \includegraphics[width=1\linewidth]{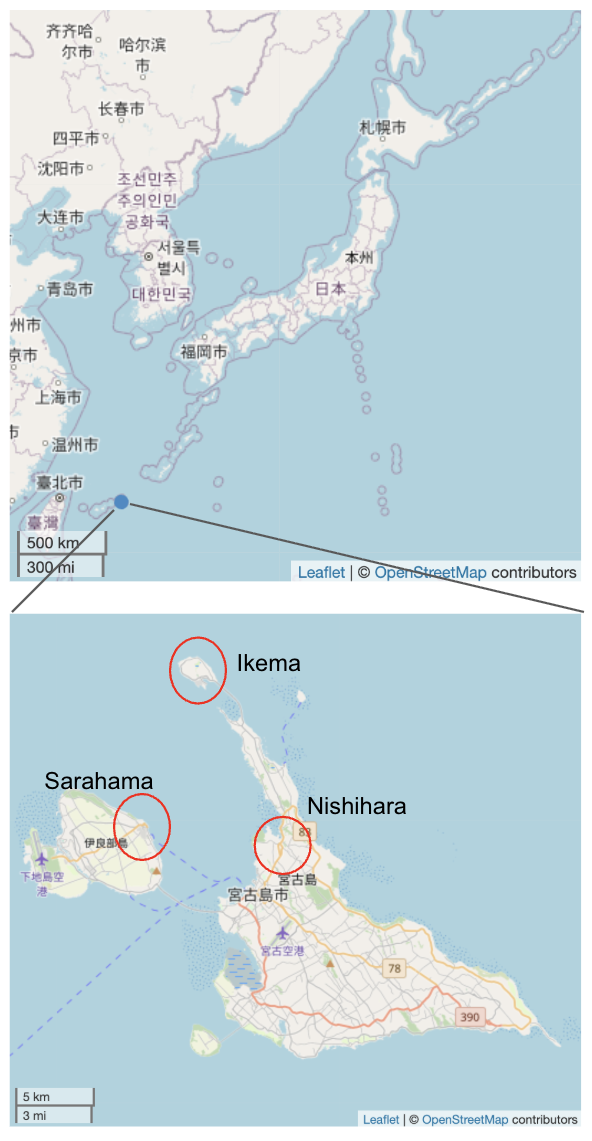}
    \caption{Map of the Miyako Islands (bottom) in Japan (top).
    The Ikema-speaking villages are marked with a red circle.}
    \label{fig:map}
\end{figure}

Ikema is an endangered Japonic language spoken in the Miyako Islands of Okinawa, Japan.
Its linguistic classification is illustrated in Figure~\ref{fig:classification}.
The language is spoken in three villages: Ikema Island, the Nishihara village on Miyako Island, and the Sarahama village on Irabu Island, as shown in Figure~\ref{fig:map}.
A recent study predicts that Ikema will become ``critically endangered'' according to UNESCO's criteria within 30 years, when the youngest active speakers reach 90 years of age \cite{nakama-2025-ikema}.
Nevertheless, Ikema is comparatively better studied and documented than other Miyakoan varieties, with an existing descriptive grammar \citep{hayashi-2013} and dictionary \citep{nakama-2025-ikema}.
\citet{YamadaEtAl2020RyukyuanIntelligibility} report that Ikema is largely unintelligible to speakers of Japonic languages outside the Southern Ryukyuan subgroup.

\input{fig/classification}

Although Ikema currently lacks an official orthography, several writing systems have been proposed to document the language.
In linguistic literature on the Ikema grammar, phonemic transcription using either the International Phonetic Alphabet (IPA) or romanization has been commonly employed.
In contrast, the speaker community more frequently uses \textit{hiragana} or \textit{katakana} (collectively referred to as \textit{kana}) which are syllabic writing systems used in Japanese \citep{takubo-2021-hougen}.
The existing dictionary \citep{nakama-2025-ikema} defines a phonemic kana orthography for Ikema, incorporating several extensions to represent phonemes absent in Modern Standard Japanese, such as a syllabic devoiced nasal consonant.
The phonemic \textit{kana} orthography can be deterministically converted into romanized script.
Example~(\ref{ex:orthography}) shows a comparison of the three writing systems: (\ref{ex:kana}) for \textit{kana}, (\ref{ex:romaji}) for phonemic romanization (\textit{romaji} henceforth), and (\ref{ex:ipa}) for phonemic IPA.
For reference, (\ref{ex:gloss})\footnote{
    \textsc{2sg}: 2nd person singular pronoun;
    \textsc{acc}: accusative case;
    \textsc{foc}: focus marker;
    \textsc{prog}: progressive aspect;
    \textsc{pst}: past tense;
    \textsc{q}: interrogative;
    \textsc{top}: topic marker
} provides the gloss and the translation of the example.

\pex\label{ex:orthography}
\a\label{ex:kana} \ja{\small っう゛ぁー~~んﾟぬ~~なうゆどぅ~~ほぅーたー}
\a\label{ex:romaji} vvaa \textsubring{N}nu nauyudu huutaa
\a\label{ex:ipa} vvaa \textsubring{n}nu naujudu huutaa
\a\label{ex:gloss}
    \begingl
        \gla vva-a \textsubring{n}nu nau-ju-du huutaa //
        \glb \textsc{2sg-top} yesterday what-\textsc{acc-foc} do.\textsc{prog.pst.q} //
        \glft `What were you doing yesterday?' //
    \endgl
\xe

\subsection{Speech recognition}
Speech recognition technologies have advanced drastically with the development of deep learning, and their applications have increasingly extended to low-resource and endangered languages.
Such efforts have led to the development of ASR systems for Nasal \citep{billings-mcdonnell-2025-connecting}, Newar and Dzardzongke \citep{oneill-2023-language}, Sardinian \citep{chizzoni-vietti-2024-towards}, Mvskoke \citep{mainzinger-levow-2024-fine}, Kichwa \citep{taguchi-etal-2024-killkan}, Japhug \citep{guillaume-etal-2022-fine}, and Khinalug \citep{li-etal-2024-speech}, among others.
Notably, several of these studies emphasize the use of ASR systems to accelerate the transcription process, which is an essential yet labor-intensive stage of language documentation.
Transcription requires significant time investment and specialized linguistic literacy and is often referred to as the ``transcription bottleneck'' in language documentation \citep{seifart-2018-language}.
Since documenting endangered languages is inherently time-sensitive, mitigating this bottleneck through ASR support represents a promising direction.
Indeed, case studies on Yongning Na \citep{michaud-2018-integrating} and Seneca \citep{jimerson-prudhommeaux-2018-asr} have demonstrated the utility of ASR integration in transcription workflows, where ASR outputs serve as drafts for human post-editing.
While these studies confirm the usefulness of the approach, quantitative evidence on the extent to which ASR accelerates transcription in practice remains limited.

%% file: fig/classification.tex
\begin{figure*}
    \centering
    \begin{forest}
        [Japonic, s sep=20mm
            [Japanese]
            [Ryukyuan, s sep=20mm
                [Southern Ryukyuan, s sep=20mm
                    [Macro--Yaeyama]
                    [Miyako
                        [Miyako]
                        [Tarama]
                        [{\=O}gami]
                        [Irabu]
                        [\textbf{Ikema}]
                    ]
                ]
                [Northern Ryukyuan
                    [Amami]
                    [Okinawa]
                ]
            ]
        ]
    \end{forest}
    \caption{A classification of Japonic languages and Ikema's position thereof.
    The classification is largely based on \citet{shimoji-2008-grammar-irabu} and \citet{pellard-2015-linguistic}.}
    \label{fig:classification}
\end{figure*}

%% file: src/3_dataset.tex
\section{Dataset}
The dataset constructed in this study is composed of three sources: video recordings collected through nearly twenty years of fieldwork (hereafter, ``Field'' data), pronounced entries from the Ikema dictionary \citep{nakama-2025-ikema} (hereafter, ``Dictionary'' data), and audiobooks.
A large portion of the Field data consists of semi-spontaneous monologues by a single male speaker.
Most of these monologues are planned speech, in which the speaker was either provided with a specific topic or has read a prepared script before recording.
The Dictionary data contain scripted speech produced by the same male speaker.
Most audio segments in the Dictionary data are shorter than one second, as each recording corresponds to a single dictionary entry (i.e., a word).
The audiobook data consist of stories read aloud by a female speaker.
See Table~\ref{tab:dataset} for the detailed statistics of the data.

\begin{table}[t]
    \centering
    \setlength{\tabcolsep}{4pt}
    \begin{tabular}{@{}lrrr@{}} \toprule
        Dataset & \#Samples & Duration & Avg. duration \\
        & & (sec) & (sec) \\ \midrule
        Field & 11126 & 16058 & 1.44 \\
        Dictionary & 5680 & 5379 & 0.95 \\
        Audiobooks & 285 & 1348 & 4.73 \\ \midrule
        Total & 17091 & 22785 & 1.33 \\
        \bottomrule
    \end{tabular}
    \caption{Statistics of the datasets.}
    \label{tab:dataset}
\end{table}

The video recordings were first converted to WAV files, after which the audio was segmented and annotated with transcriptions.
Transcriptions were provided in \textit{kana} in a phonetically faithful manner; that is, utterances were transcribed as they were pronounced, including fillers, disfluencies, and repetitions.
After the kana transcription was completed, it was transliterated into \textit{romaji} using deterministic mapping rules.
In the dataset, word boundaries are indicated by a half-width space (U+0009).

\begin{figure*}
    \centering
    \includegraphics[width=1\linewidth]{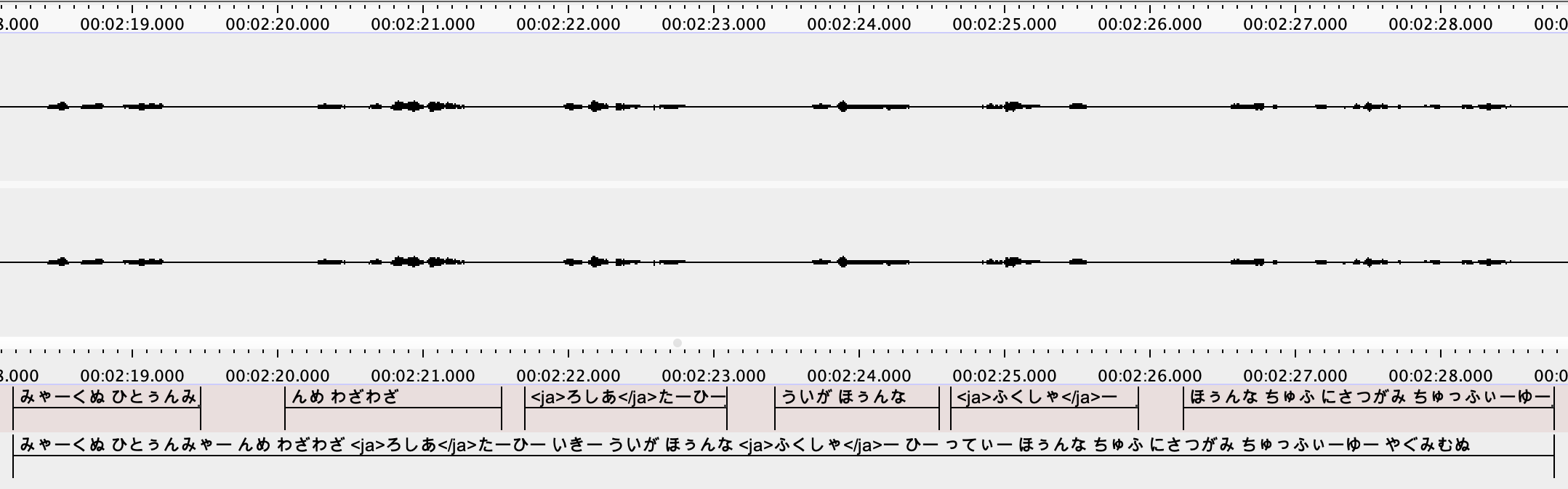}
    \caption{Segmentation and annotation in ELAN.
    The top transcription tier contains pause-based segments, while the bottom tier contains longer segments that concatenate multiple pause-based segments to approximate sentence-level units.}
    \label{fig:ELAN}
\end{figure*}

The transcriptions include several types of tags that provide detailed linguistic information, such as code-switched segments in Japanese (\texttt{<ja>...</ja>}), disfluencies (\texttt{<dis>...</dis>}), songs (\texttt{<song>...</song>}), and personal names (\texttt{<name>...</name>}).\
In addition, utterances that annotators were unable to understand are annotated with the tag \texttt{<unsure>...</unsure>}.
This design allows users who wish to train models on clean transcriptions without disfluencies or repetitions to automatically remove them from the dataset, and also enables the easy exclusion of samples containing personally identifiable information.
Given the nature of spontaneous speech, sentence boundaries are often difficult to determine.
Therefore, segment boundaries were determined based on pauses between speech regions.
In addition, the annotated data include a tier with longer segments that roughly correspond to sentence-level units, as shown in Figure~\ref{fig:ELAN}.

%% file: src/4_experiments.tex
\section{Experiments}
\subsection{Setup}

In our experiments, we train automatic speech recognition (ASR) models on the newly developed Ikema speech dataset.
Specifically, we fine-tune pretrained Wav2Vec2 models \citep{baevski2020wav2vec20frameworkselfsupervised} using a Connectionist Temporal Classification (CTC) decoder layer \citep{graves-2006-connectionist}.
Wav2Vec2 is a self-supervised model that learns speech representations from large amounts of unlabeled audio.
Among its multilingual variants, XLS-R \citep{babu2021xlsrselfsupervisedcrosslingualspeech} and MMS \citep{pratap2023scalingspeechtechnology1000} have been trained on 128 and 1,406 languages, respectively, enabling robust cross-lingual transfer.

For Ikema, we fine-tune these pretrained models by adding a final CTC layer that maps frame-level features to textual transcriptions. The CTC layer produces a token prediction for each frame, and identical consecutive tokens are collapsed into a single symbol to form the decoded output, typically at the character or grapheme level.
Wav2Vec2-based models have been shown to perform effectively even in extremely low-resource settings \citep{liang-levow-2025-breaking}, making them well-suited for our experiments.

While automatic speech recognition (ASR) models with autoregressive decoders, such as Whisper \citep{radford2022robustspeechrecognitionlargescale}, achieve strong performance for high-resource languages, they require large amounts of data to train the decoder to model subword dependencies effectively.
In contrast, a model with a standard CTC decoder does not explicitly learn dependencies across tokens, allowing it to directly learn the alignment between the input speech and the corresponding output characters at each time frame.
Furthermore, CTC's per-frame outputs are particularly suitable for our goal of producing phonetically faithful transcriptions, as the transcribed data may contain substantial disfluencies and fillers that should not be omitted in language documentation.

In our experiments, we compare the performance of \texttt{wav2vec2-xls-r-300m} (300M parameters, 128 languages), \texttt{wav2vec2-xls-r-1b} (1B parameters, 128 languages), and \texttt{mms-1b} (1B parameters, 1,406 languages).
The tags in the transcription are removed in the preprocessing.
Of the dataset,
80\% is used as the training data, 10\% is reserved for the validation data, and 10\% for the test data.
We also train models in two writing systems: \textit{kana} and \textit{romaji}.
The token vocabulary is constructed based on syllables (or more precisely, morae) for \textit{kana} and phonemes for \textit{romaji}; for example, ``\ja{じゃ}'' /{\textctz}a/ is counted as one token in the \textit{kana} model, and ``zy'' /\textctz/ as one token in the \textit{romaji} model.

We report character error rate (CER) and word error rate (WER), defined as the percentage of incorrect characters or words out of 100.
It is important to note, however, that WER is less informative for Ikema, where word boundaries are not well defined and variations appear even in the gold transcriptions.

The learning rate is set to 0.0003.
Preliminary experiments showed higher learning rates exhibited to slower and unstable learning curves.
The batch size was set to 16 with \texttt{wav2vec2-xls-r-300m} and 4 with \texttt{wav2vec2-xls-r-1b} and \texttt{mms-1b}.
The model is trained for 50 epochs.
All the models are trained on a single A10 GPU with 16GB RAM.

\subsection{Results}
Table~\ref{tab:results} shows the performance of the trained ASR models under different conditions.
Compared to the \textit{kana} transcription, the \textit{romaji} transcription consistently yielded lower CERs.
This is likely because the \textit{kana} writing system encodes more phological information than \textit{romaji}.
This finding aligns with the general tendency of Wav2Vec2 models to underperform when transcribing more complex writing systems \citep{taguchi-chiang-2024-language}.

\begin{table}[t]
    \centering
    \begin{tabular}{lrrrr} \toprule
        & \multicolumn{2}{c}{\textit{kana}} & \multicolumn{2}{c}{\textit{romaji}} \\
        Model &                 CER &   WER &   CER &   WER \\
        \midrule
        \texttt{xls-r-300m} &   \textbf{20.94} & \textbf{62.56} & \textbf{14.80} & \textbf{64.99} \\
        \texttt{xls-r-1b} &     24.07 & 66.61 & 20.33 & 76.73 \\
        \texttt{mms-1b} &       21.86 & 63.22 & 18.98 & 73.87 \\
        \bottomrule
    \end{tabular}
    \caption{ASR performances of the trained models.}
    \label{tab:results}
\end{table}

\begin{figure}
    \centering
    \includegraphics[width=\linewidth]{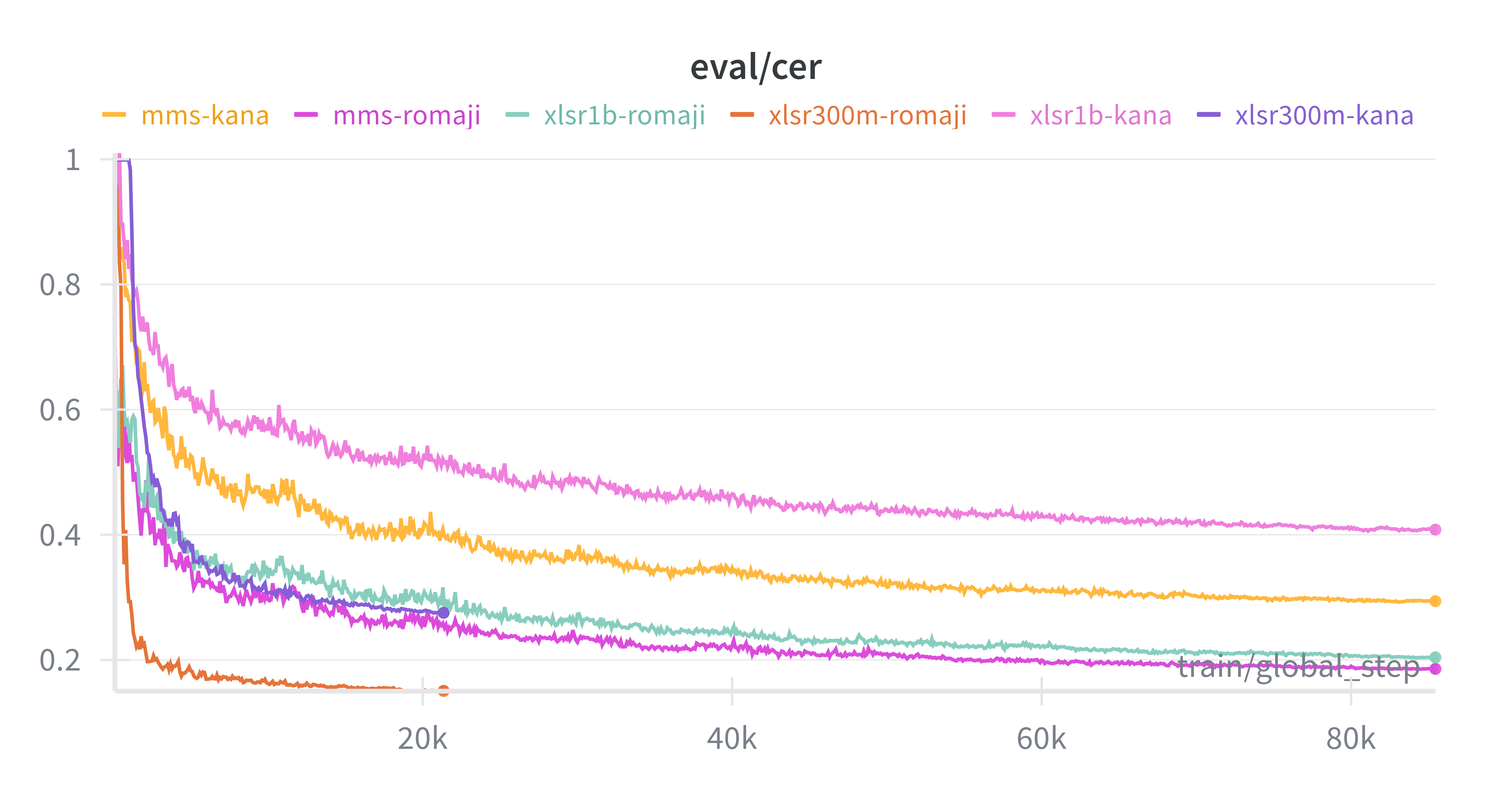}
    \caption{The curves of the CERs on the validation data during the training.}
    \label{fig:evalcer}
\end{figure}

The best performing model was the fine-tuned model based on \texttt{xls-r-300m} in both \textit{kana} and \textit{romaji} transcriptions.
On the other hand, the larger model (\texttt{xls-r-1b}) performed the worst among the three models.
These results suggest that a larger model size does not necessarily promise a successful training.
The lightest model (\texttt{xls-r-300m}) was not only effective in its performance but also advantageous in the training time and the total training steps because of its increased batch size as illustrated in Figure~\ref{fig:evalcer}.
In the experiments, the training with \texttt{xls-r-300m} completed in approximately 10 hours, while the other 1B models took more than 50 hours to complete.
In addition, the lightweight model takes up to approximately 1.2GB as opposed to the 1B models' 3.8GB, and is beneficial for running locally in the integrated annotation tool.

%% file: src/5_analysis.tex
\section{Is ASR-assisted transcription helpful?}
Whether ASR can truly benefit annotators in the transcription process has been a point of debate.
Although many studies have argued for the potential advantages of ASR in language documentation, \citet{prudhommeaux-2021-automatic} reported that members of some speaker communities preferred unassisted transcription without ASR support.
To examine this question in the case of Ikema, we conducted a human evaluation to assess the extent to which ASR can actually accelerate the transcription process.

\subsection{Setup}
For this experiment, the fine-tuned \texttt{wav2vec2-xlr-s-300m} model trained on the \textit{kana}-script is integrated into ELAN, linguistic annotation software \cite{wittenburg-etal-2006-elan}, as an extension to transcribe Ikema automatically on the user interface.
In this experiment, we choose an unannotated 5-minute recording, segment the spoken parts, and asked two annotators to annotate the same segments on ELAN under the following conditions.
The total number of segments is 121, and their total duration is 217.64 seconds.
Annotator A is a non-native expert who has worked on the documentation of Ikema for 19 years, and Annotator B is a non-native learner with less than one year of experience.
Both are native speakers of Japanese and familiar with the \textit{kana} orthography of Ikema.
Annotator A is given an EAF file  (an extended XML for ELAN containing the annotation text) in which the first half of the segments (61 samples, 110.12 seconds) are blank and the second half (60 samples, 107.52 second) are filled in with the model's prediction output.
Conversely, Annotator B is provided with an EAF file where the first half contains the model's transcription and the second half is left blank.
Both annotators are asked to transcribe the audio manually, and their annotation times are recorded separately for each half.
Note that they are both allowed to have listened to the recording before and know what is talked about.

\subsection{Results}

\begin{table}[tb]
    \centering
    \begin{tabular}{lrrr} \toprule
         & w/o ASR & w/ ASR & Speedup\\ 
        \midrule
        Annotator A & 11:36 & 9:43 & +19.38\% \\
        Annotator B & 30:49 & 25:00 & +23.27\% \\
        \bottomrule
    \end{tabular}
    \caption{A comparison of annotation speed, with and without ASR output.
    The speedup is computed by dividing the time without ASR by the time with ASR.}
    \label{tab:time}
\end{table}

As Table~\ref{tab:time} demonstrates, both annotators reported a positive speedup in ASR-assisted transcription.
The annotators also noted that they felt it somewhat easier to annotate the data with the ASR draft despite its errors.
Annotator B noted that segments with a frequently occurring filler and a fixed expression were often transcribed correctly, decreasing the need of manual correction with clicking, typing, and re-listening.
The results confirmed the positive effect of ASR outputs in language documentation even when the model's accuracy is not ideal.

As for the inter-annotator agreement between the two annotators, there was a CER of 17.32\% and a WER of 46.24\%.
The CER of the ASR output was 29.61\% with Annotator A, and 25.25\% with Annotator B.
However, when minor orthographic ambiguities are ignored, the CER of the ASR output decreases to 20.34\%.
The high error rates reflect the non-standardized aspects of the language's writing system, such as post-lexical morphophonological changes \cite{Takubo2021IkemaMorphophonemics} and the position of whitespaces.
Example~(\ref{ex:eval}) shows an instance of such cases.\footnote{
    \textsc{cond}: conditional conjunctive suffix;
    \textsc{dsc}: discourse sentence-final particle;
    \textsc{inf}: infinitive form.
}
The words \textit{umatti} `coming here' and \textit{Nmeera} `umm' (a filler) are often pronounced as if they are single words.
For this reason, Annotator A (\ref{ex:eval-a}) transcribed them as single words, while Annotator B (\ref{ex:eval-b}) inserted spaces based on their underlying analyzed forms. 
In addition, while Annotator A often omitted disfluencies and repetitions, Annotator B transcribed the speech faithfully including the speech errors.

For reference, (\ref{ex:eval-asr}) shows the actual output by the recognizer.
While the high error rates may suggest low quality, a closer look at the output reveals that the model's prediction reflects how the sample sounds.
In fact, the suffix \textit{-tigaa} often realizes as phonetically shortened forms \textit{-tiyaa}, \textit{-tiya}, or \textit{-taa}, the filler \textit{Nme} is also pronounced as \textit{Nmya}, and the sentence-final particle \textit{ira} can also realize as \textit{ra}.
This phonetic faithfulness is characteristic of the CTC decoder outputting frame by frame without a language model.
While the model still mispredicts token boundaries, the errors observed in the outputs are often in fact free variations.
This observation is coherent with the discussion provided in the previous study on ASR for Japhug \cite{guillaume-etal-2022-fine}.

\lingset{labeloffset=0em, textoffset=0em, interpartskip=0.1em}
\pex[nopreamble=true]\label{ex:eval}
\a\label{ex:eval-a}
    \begingl[belowglpreambleskip=0ex, aboveglftskip=0ex]
        \glpreamble \ja{\small うまってぃ~~うぐなーいてぃがー~~んめーら} //
        \gla umatti ugunaaitigaa Nmeera //
        \glb {\it uma+tti} {\it ugunaai-tigaa} {\it Nme+ira} //
        \glc here+come.\textsc{inf} gather-\textsc{cond} well+\textsc{dsc} //
        \glft `If (they) come here and get together, well' //
    \endgl
\a\label{ex:eval-b}
    \begingl[belowglpreambleskip=0ex]
        \glpreamble \ja{\small うま~~ってぃ~~うぐなーいてぃがー~~んめ~~いら} //
        \gla uma tti ugunaaitigaa Nme ira //
        \glb {\it uma} {\it tti} {\it ugunaai-tigaa} {\it Nme} {\it ira} //
        \glc here come.\textsc{inf} gather-\textsc{cond} well \textsc{dsc} //
    \endgl
\a\label{ex:eval-asr}
    \begingl[belowglpreambleskip=0ex]
        \glpreamble \ja{\small うま~~てぃうぐないたーんみゃら} //
        \gla uma tiugunaitaaNmyara //
    \endgl
\xe

%% file: src/6_conclusion.tex
\section{Conclusion}

This study presented an ongoing effort to develop an automatic speech recognition (ASR) system for Ikema Miyakoan, an endangered Ryukyuan language spoken in Okinawa, Japan.
We constructed a {\totaldatasethours}-hour speech corpus from recordings through collaborative fieldwork with the speaker community.
Based on this dataset, we trained an ASR model achieving a CER as low as {\bestromajicer}\%.
Then, we integrated the model into ELAN as an ASR extension and evaluated its potential to mitigate the transcription bottleneck in language documentation.
The results indicate that the integration of ASR into transcription workflows can substantially accelerate annotation and reduce the cognitive burden on human transcribers.

%% file: table/dataset.tex
\begin{table*}[t]
    \centering
    \begin{tabular}{llrrrr} \toprule
        Recording ID & Style & Duration & \#Samples & \#Kana & \#Words \\
        \midrule
        I0482\_usI & Spontaneous & 111.03 & 69 & 1105 & 202 \\
        I0482\_ngi & Spontaneous & 100.36 & 72 & 885 & 201 \\
        I0482\_pic\_mucIusa & Spontaneous & 85.93 & 56 & 683 & 132 \\
        I0482\_pic\_buugii & Spontaneous & 72.91 & 49 & 534 & 117 \\
        I0482\_byuuigassa & Spontaneous & 53.51 & 35 & 454 & 98 \\
        I0482\_bippii & Spontaneous & 26.27 & 15 & 229 & 43 \\
        I0482\_pic\_barazan & Spontaneous & 30.50 & 16 & 324 & 58 \\
        I0482\_niguu & Spontaneous & 86.81 & 56 & 675 & 148 \\
        I0482\_pic\_takanna & Spontaneous & 44.99 & 34 & 412 & 83 \\
        I0482\_pic\_taka & Spontaneous & 179.77 & 108 & 1698 & 345 \\
        I0412\_suuni & Spontaneous & 29.76 & 21 & 246 & 48 \\
        I0483\_0414\_satatinpura & Spontaneous & 44.30 & 29 & 382 & 77 \\
        I0482\_pic\_zzakugii & Spontaneous & 73.37 & 48 & 586 & 122 \\
        I0482\_gisIcI & Spontaneous & 107.87 & 72 & 934 & 193 \\
        I0482\_kami & Spontaneous & 197.85 & 138 & 1521 & 332 \\
        I0482\_gazIhanagii & Spontaneous & 47.87 & 29 & 426 & 86 \\
        I0482\_ugan & Spontaneous & 84.47 & 61 & 677 & 129 \\
        I0483\_0277\_yasaiitame & Spontaneous & 19.33 & 16 & 114 & 27 \\
        I0483\_0276\_waa & Spontaneous & 16.76 & 10 & 143 & 30 \\
        I0483\_0405\_isIusI & Spontaneous & 114.51 & 77 & 942 & 203 \\
        I0483\_0527\_suba & Spontaneous & 37.64 & 22 & 345 & 69 \\
        I0483\_0412\_waanimun & Spontaneous & 34.99 & 26 & 283 & 61 \\
        I0483\_0688\_mamisuimai & Spontaneous & 63.79 & 40 & 540 & 110 \\
        I0483\_0409\_fukyagi & Spontaneous & 66.88 & 45 & 568 & 111 \\
        I0483\_dakyau & Spontaneous & 19.40 & 15 & 127 & 27 \\
        I0482\_pc\_taummabasI & Spontaneous & 324.87 & 185 & 3049 & 586 \\
        I0483\_0678\_saguna & Spontaneous & 118.40 & 70 & 1093 & 220 \\
        I0482\_kkucIgii\ & Spontaneous & 65.86 & 46 & 534 & 111 \\
        I0483\_0222\_avvansu & Spontaneous & 23.82 & 19 & 180 & 39 \\
        I0488\_\ja{しーさー} & Spontaneous & 100.76 & 69 & 832 & 180 \\
        I0490\_\ja{あまん} & Spontaneous & 111.11 & 79 & 958 & 214 \\
        I0490\_\ja{池間大橋} & Spontaneous & 167.68 & 113 & 1485 & 306 \\
        I0496\_kaichosyokureki & Conversation & 912.26 & 339 & 8407 & 1706 \\
        I0501\_nakasonetuimyaa & Spontaneous & 305.98 & 86 & 2495 & 461 \\
        I0502\_harimizuutaki & Spontaneous & 405.51 & 151 & 3419 & 660 \\
        I0503\_Nevskynohi & Spontaneous & 145.16 & 60 & 1054 & 204 \\
        I0506\_aisatsutehon\_Va & Spontaneous & 19.18 & 7 & 85 & 19 \\
        I0506\_aisatsutehon\_Vb & Spontaneous & 13.60 & 6 & 62 & 19 \\
        I0506\_aisatsutehon\_Vc & Spontaneous & 15.88 & 7 & 109 & 21 \\
        I0506\_aisatsutehon\_Vd & Spontaneous & 101.15 & 25 & 639 & 131 \\
        I0506\_aisatsutehon\_Ve & Spontaneous & 159.49 & 47 & 1260 & 247 \\
        I0506\_aisatsutehon\_Vf & Spontaneous & 79.75 & 17 & 612 & 122 \\
        I0507\_mingukaisetsu\_Va & Spontaneous & 112.81 & 67 & 978 & 215 \\
        I0507\_mingukaisetsu\_Vb & Spontaneous & 126.18 & 39 & 942 & 191 \\
        I0508\_nakamautaki\_V & Spontaneous & 367.59 & 102 & 2716 & 507 \\
        I0509\_zyaagama\_V & Spontaneous & 254.00 & 76 & 1918 & 347 \\
        I0510\_kyuukoominkan\_Vb & Spontaneous & 116.36 & 22 & 787 & 152 \\
        I0510\_kyuukoominkan\_Va & Spontaneous & 182.85 & 39 & 1387 & 263 \\
        \bottomrule
    \end{tabular}
    \caption{The detail of the spontaneous speech dataset (Field). The Audiobook data and the Dictionary data are not shown in this table.}
    \label{tab:dataset-full}
\end{table*}